\documentclass[letterpaper, 10 pt, conference]{ieeeconf}  

\IEEEoverridecommandlockouts                              

\usepackage{graphicx} 
\usepackage{color} 
\usepackage{amsmath} 
\usepackage{amssymb}  
\usepackage{makecell} 
\usepackage{multirow}
\usepackage{subcaption}
\usepackage[linesnumbered,ruled,vlined]{algorithm2e}
\usepackage[colorlinks, linkcolor=blue]{hyperref}

\usepackage{epstopdf}
\AppendGraphicsExtensions{.tif}

\title{\LARGE \bf
Task-Oriented Dexterous Hand Pose Synthesis \\Using Differentiable Grasp Wrench Boundary Estimator
}

\author{Jiayi Chen$^{1, 2}$, Yuxing Chen$^{1,3}$, Jialiang Zhang$^{1,3}$ and He Wang$^{1, 2, 3\dagger}$
\thanks{$^{1}$CFCS, School of Computer Science, Peking University.}%
\thanks{$^{2}$Beijing Academy of Artificial Intelligence.}%
\thanks{$^{3}$Galbot.}%
\thanks{Corresponding author: hewang@pku.edu.cn}
}

\newtheorem{property}{Property}

\begin{document}

\newcommand{\jiayi}[1]{{\color{red}[Jiayi: #1]}}
\newcommand{\yuxing}[1]{{\color{blue}[Yuxing: #1]}}
\newcommand{\jialiang}[1]{{\color{green}[Jialiang: #1]}}
\newcommand{\he}[1]{{\color{cyan}[He: #1]}}
\makeatletter
\newcommand{\newparallel}{\mathrel{\mathpalette\new@parallel\relax}}
\newcommand{\new@parallel}[2]{%
  \begingroup
  \sbox\z@{$#1T$}
  \resizebox{!}{\ht\z@}{\raisebox{\depth}{$\m@th#1/\mkern-5mu/$}}%
  \endgroup
}
\makeatother

\definecolor{TWSblue}{rgb}{0.098,0.482,0.725}
\definecolor{GWSgreen}{rgb}{0.492,0.894,0.749}

\maketitle
\thispagestyle{empty}
\pagestyle{empty}

%

\begin{abstract}

This work tackles the problem of task-oriented dexterous hand pose synthesis, which involves generating a static hand pose capable of applying a task-specific set of wrenches to manipulate objects. 
Unlike previous approaches that focus solely on force-closure grasps, which are unsuitable for non-prehensile manipulation tasks (\textit{e.g.}, turning a knob or pressing a button), we introduce a unified framework covering force-closure grasps, non-force-closure grasps, and a variety of non-prehensile poses.
Our key idea is a novel optimization objective quantifying the disparity between the Task Wrench Space (TWS, the desired wrenches predefined as a task prior) and the Grasp Wrench Space (GWS, the achievable wrenches computed from the current hand pose). By minimizing this objective, gradient-based optimization algorithms can synthesize task-oriented hand poses without additional human demonstrations.
Our specific contributions include 1) a fast, accurate, and differentiable technique for estimating the GWS boundary; 2) a task-oriented objective function based on the disparity between the estimated GWS boundary and the provided TWS boundary; and 3) an efficient implementation of the synthesis pipeline that leverages CUDA accelerations and supports large-scale paralleling.
Experimental results on 10 diverse tasks demonstrate a 72.6\% success rate in simulation. Furthermore, real-world validation for 4 tasks confirms the effectiveness of synthesized poses for manipulation. Notably, despite being primarily tailored for task-oriented hand pose synthesis, our pipeline can generate force-closure grasps 50 times faster than DexGraspNet while maintaining comparable grasp quality.
Project page: \href{https://pku-epic.github.io/TaskDexGrasp/}{https://pku-epic.github.io/TaskDexGrasp/}.

\end{abstract}

\section{Introduction}


Robotic dexterous hands hold great promise for anthropomorphic manipulation due to their humanoid structure. 
In recent years, vision-based dexterous grasping methods~\cite{xu2023unidexgrasp,li2023gendexgrasp,lu2023ugg} have seen substantial improvements in generalizability. 
These successes are empowered by large-scale grasping datasets~\cite{wang2023dexgraspnet,turpin2023fast} generated by grasp synthesis algorithms~\cite{liu2021synthesizing,turpin2022grasp,li2023frogger}. 
However, these synthesis algorithms are all limited to generating force-closure grasps, \textit{i.e.}, grasp poses capable of resisting external wrenches from any direction. Nonetheless, synthesizing hand poses for non-force-closure grasping and non-prehensile manipulation is also crucial, especially when a force-closure grasp is unnecessary or unattainable, such as turning a knob or pressing a button.

\begin{figure}[t]
    \centering
    \includegraphics[width=\linewidth]{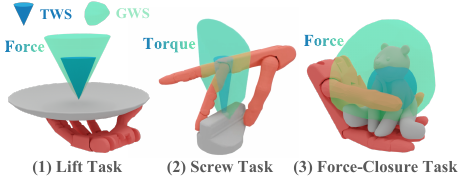}
    \caption{\textbf{Task-oriented dexterous hand poses synthesized for 3 tasks.} Each {\color{TWSblue}TWS} is specified by humans as a task prior: (1) the lift task requires upper forces, (2) the screw task requires counter-clockwise torques, and (3) the force-closure task requires wrenches in all directions. The {\color{GWSgreen}GWS} is computed based on the contacts between the hand and the object. 
    Both {\color{TWSblue}TWS} and {\color{GWSgreen}GWS} are visualized by their 3D force or torque components. }
    \label{fig:teaser}
\end{figure}

To address the aforementioned limitation, we propose a unified framework to synthesize dexterous hand poses for various tasks, including force-closure grasping, non-force-closure grasping, and several typical non-prehensile manipulations. 
Our key insight is to align the Task Wrench Space (TWS) and the Grasp Wrench Space (GWS), where TWS represents the wrench set that the hand \textit{should} apply to the object and GWS represents what the hand \textit{can} apply.
Formally, we achieve this by constructing an objective function and minimizing it using gradient-based optimization. 
This approach can automatically synthesize task-oriented hand poses by simply providing an object mesh and defining a TWS, without requiring additional human demonstrations. Some examples are shown in Fig. \ref{fig:teaser}.

Specifically, our first contribution is a fast, accurate, and differentiable algorithm to estimate the GWS boundary under the max-magnitude ($L_\infty$) bound assumption~\cite{krug2017grasp}, \textit{i.e.} each contact force is restricted to a maximum value of 1. 
Our key insight is that, by leveraging the GWS's specific mathematical properties, we can construct a surjection from 6D unit vectors to its surface, thereby facilitating dense sampling to estimate the GWS boundary (GWB).
The computation scales linearly with the contact number $m$ and sample number $K$, and the algorithm is parallelizable with respect to the contacts and samples, enabling scalability in contact-rich scenarios.






Second, we propose a novel task-oriented objective function for gradient-based optimization. 
This function measures the disparity between the given TWS and the estimated current GWS, thereby encouraging the hand pose to gradually conform to the task specifications during optimization. 
Formally, we represent the TWS as a 6D hyper-spherical sector parametrized by a 6D unit vector and an angle. In particular, when this angle equals $180^\circ$, the TWS degenerates to a 6D hypersphere, indicating the force-closure scenario. 
This formulation of the TWS provides operators with a convenient means to specify task objectives. Nevertheless, our objective function is not limited to this representation.

Finally, we implement the pose synthesis pipeline using cuRobo~\cite{curobo_report23}, a motion planning library based on trajectory optimization designed to run on GPUs. Leveraging their CUDA kernels for parallelized forward kinematics, collision checking, and numerical optimization, our pipeline is highly efficient and supports large-scale paralleling.

Experimental results verify the efficiency and effectiveness of our methods. Our GWS boundary estimator can run in milliseconds on GPU, $200\times$ faster than the traditional discretization-based method. Moreover, we design 10 different tasks and achieve an overall success rate of $72.6\%$ in simulation. Real-world validation on 4 tasks further confirms the synthesized poses for manipulation. Notably, despite being primarily
tailored for task-oriented hand pose synthesis, our pipeline can synthesize 100,000 force-closure grasps for 5,000 objects within 1.2 GPU hours, $50\times$ faster than DexGraspNet~\cite{wang2023dexgraspnet}, with comparable grasp quality.

In summary, our main contributions are (i) a fast, accurate, and differentiable algorithm to estimate the Grasp Wrench Space boundary, (ii) task-oriented objective function for gradient-based hand pose synthesis, and (iii) a heavily optimized dexterous hand pose synthesis pipeline that runs 50 times faster than DexGraspNet.

\section{Related Work}

\textbf{Task-Oriented Grasp Analysis.} TWS is frequently used to describe the wrenches needed for grasping and manipulating objects. Previous studies~\cite{li1988task, el2015computing, lin2015grasp, borst2004grasp} approximate TWS as either a 6D ellipsoid or a sphere to simplify grasp analysis, but these formulations are limited to force-closure grasps. Only a few studies~\cite{kruger2011partial, kruger2012local} have looked into non-force-closure cases, typically focusing on using two or three frictionless fingers to grasp objects with polygonal shapes. Our work defines TWS as a 6D hyper-spherical sector, which allows us to handle both force-closure and non-force-closure scenarios. Moreover, our method can accommodate dexterous hands with four or five fingers and complex object meshes.

\textbf{Grasp Wrench Space Estimation.}
GWS is the basis for most grasp quality metrics~\cite{roa2015grasp}. Estimating GWS commonly involves calculating the convex hull over discretized friction cones assuming a sum-magnitude ($L_1$) constraint, where all contact forces sum up to 1~\cite{ferrari1992planning}. However, this assumption can be overly conservative and thus unsuitable for multi-finger dexterous hands~\cite{krug2017grasp}. Conversely, under the $L_\infty$ assumption, the conventional discretization-based method exhibits exponential time complexity with respect to the number of contacts~\cite{ferrari1992planning}. Some studies~\cite{zheng2009improving, zheng2012efficient, borst2004grasp} have utilized optimization techniques to accelerate computation under the $L_\infty$ assumption, but these approaches are non-differentiable with respect to hand pose. We propose an efficient and differentiable algorithm to approximate the GWS boundary under the $L_\infty$ assumption.

\textbf{Force-Closure Dexterous Grasp Synthesis.} 
Dexterous grasp synthesis involves both analytical and data-driven methods. Early reviews on this topic can be found in~\cite{sahbani2012overview, bohg2013data}. Analytical methods focus on optimizing certain quality metrics, such as the $\epsilon$ metric, to discover effective grasps. Previous approaches, like \textit{GraspIt!}~\cite{miller2004graspit}, rely on sampling-based techniques, which are inefficient for dexterous hands with a high degree of freedom. More recent studies~\cite{liu2021synthesizing, li2023frogger, dai2018synthesis} employ gradient-based optimization with differentiable approximations of the $\epsilon$ metric. However, these approximations often lack strong physical interpretations and are limited to force-closure grasps. Another approach~\cite{turpin2022grasp, turpin2023fast} synthesizes dexterous grasps using a differentiable simulator; however, the simulator's gradient may be inaccurate~\cite{zhong2022differentiable}. 

Data-driven methods~\cite{shao2020unigrasp, wei2022dvgg} rely on learning from data. They often utilize analytical methods for data preparation~\cite{wang2023dexgraspnet, turpin2023fast} and post-processing~\cite{jiang2021hand, li2023gendexgrasp} because collecting extensive human data is costly and network predictions may lack accuracy. 
Recent advancements in dexterous grasping have embraced reinforcement learning~\cite{mandikal2022dexvip, wan2023unidexgrasp++} and imitation learning~\cite{qin2022dexmv}. While RL can achieve satisfying results without synthetic datasets, its efficacy is augmented when combined with such data, as exemplified by \cite{xu2023unidexgrasp}. This emphasizes the ongoing significance of hand pose synthesis.

\textbf{Non-Prehensile Manipulation.} Non-prehensile manipulation encompasses various tasks, including pushing~\cite{arruda2017uncertainty}, catching~\cite{kim2014catching}, pivoting~\cite{chen2023synthesizing}, and in-hand manipulation~\cite{chen2023visual}. Given their significant differences from force-closure grasps, prior studies~\cite{cheng2023enhancing, pang2023global} often employ different strategies than conventional grasp synthesis algorithms. In contrast, we present a unified framework for generating both force-closure grasps and various non-prehensile hand poses, including pushing, pulling, lifting, turning, etc.

\section{Preliminaries}

Consider an object grasped by a hand with $m$ contacts. For each contact $i \in [1, 2, \cdots, m]$, let $\mathbf{p}_i\in \mathbb{R}^3$ be the contact position, $\mathbf{n}_i\in \mathbb{R}^3$ be the inward-pointing surface unit normal, and $\mathbf{d}_i,\mathbf{e}_i\in \mathbb{R}^3$ be two unit tangent vectors such that $\mathbf{n}_i = \mathbf{d}_i \times \mathbf{e}_i$, all defined in the object coordinate frame. Although our method also works for the soft contact (SFC) model (please refer to the Appendix), in our main paper, we use the point contact with friction (PCF) model:
\begin{flalign}
\label{eq: F_pcf}
& \mathcal{F}_i^{PCF} = \left\{\mathbf{f}_i\in\mathbb{R}^3~|~0 \leq f_{i1}, f_{i2}^2+f_{i3}^2 \leq \mu^2 f_{i1}^2 \right\} 
\\
\label{eq: G_pcf}
& \mathbf{G}_i^{PCF} = 
\begin{bmatrix}
    \mathbf{n}_i & \mathbf{d}_i & \mathbf{e}_i \\
    \mathbf{p}_i \times \mathbf{n}_i & 
    \mathbf{p}_i \times \mathbf{d}_i & 
    \mathbf{p}_i \times \mathbf{e}_i \\
\end{bmatrix} \in \mathbb{R}^{6\times3}
\end{flalign}
Here, $\mathcal{F}_i$ represents all potential forces generated by contact $i$, and the matrix $\mathbf{G}_i$ maps the force set $\mathcal{F}_i$ to the wrench set $\mathcal{W}_i = \mathbf{G}_i \mathcal{F}_i$. We denote $\mathbf{f}_i=[f_{i1}, f_{i2}, f_{i3}]$ in the $\langle \mathbf{n}_i, \mathbf{d}_i, \mathbf{e}_i\rangle$ coordinate frame. $\mu$ is the friction coefficient. The resulting wrench that the robot hand can apply to the object is given by $\mathbf{w} = \sum_{i=1}^{m}{\mathbf{G}_i \mathbf{f}_i}$, where $\mathbf{f}_i \in \mathcal{F}_i$. 

All possible $\mathbf{w}$ constitute the \textit{Grasp Wrench Space (GWS)}. In this work, we employ the $L_\infty$ assumption to calculate the GWS: $\mathcal{W}_g=\bigoplus_{i=1}^{m}{\mathcal{W}_i}$, representing the Minkowski sum of each ${\mathcal{W}_i}$. We prefer the $L_\infty$ assumption over the $L_1$ assumption due to its greater suitability for multi-finger dexterous hands~\cite{krug2017grasp}. \textit{Grasp Wrench Hull (GWH)} and \textit{Grasp Wrench Boundary (GWB)} are defined as the convex hull and the boundary of $\mathcal{W}_g$, respectively, denoted as $\mathcal{W}_g^{ch}$ and $\partial\mathcal{W}_g$.

With these notations, $\epsilon$ metric~\cite{ferrari1992planning} and its task-oriented extension~\cite{borst2004grasp} can be written as 
\begin{equation}
\label{eq:epsilon}
    \epsilon=\underset{\mathbf{w}\in \partial\mathcal{W}_g}{\min}\|\mathbf{w}\|,~
    \epsilon_t = \underset{\mathbf{w}\in\partial\mathcal{W}_g, \mathbf{t}\in\mathcal{W}_t,  \mathbf{w} \newparallel \mathbf{t}} \min \frac{\|\mathbf{w}\|}{\|\mathbf{t}\|}
\end{equation}
Here, $\mathcal{W}_t$ denotes the TWS, $\|\cdot\|$ represents the L2 length norm, and $\newparallel$ indicates parallelism.


\section{Method}

\subsection{Grasp Wrench Boundary Estimator}
\label{sec:gws}

\begin{figure}[t]
  \centering
  \includegraphics[width=\linewidth]{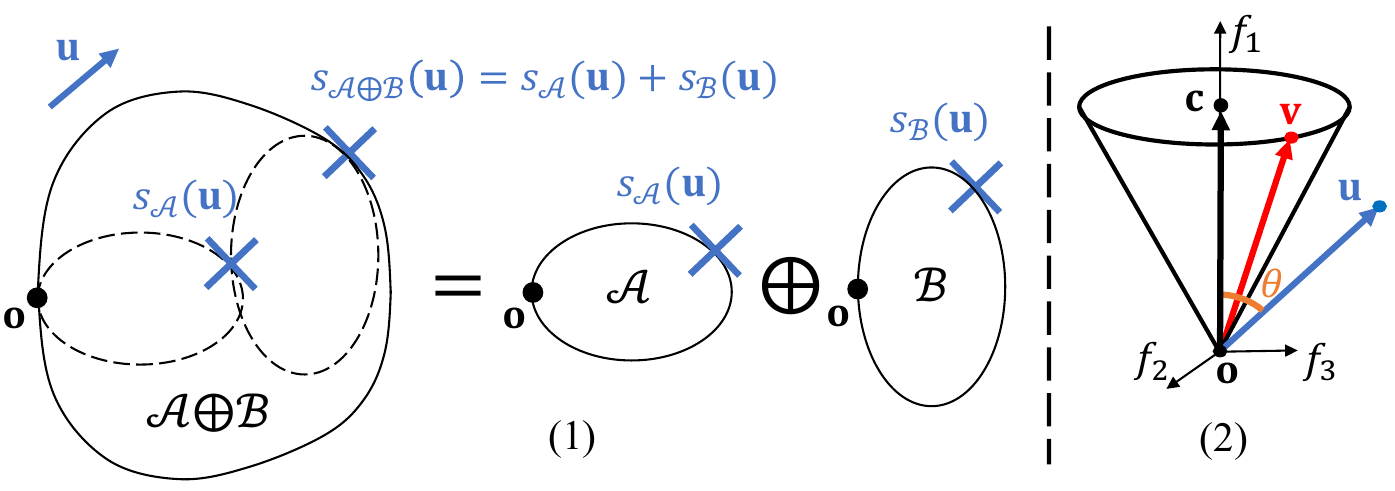}
  \caption{\textbf{Method illustration.} (1) The support mapping $s_{\mathcal{A}}(\mathbf{u})$ and its Prop. \ref{prop_add}. (2) $s_{\mathcal{F}}(\mathbf{u})$ for PCF contact model in 3D.}
  \label{fig:gws}
\end{figure}

Inspired by~\cite{zheng2009improving, qiu2022new}, we propose a novel, fast, accurate, and differentiable algorithm for constructing the GWB through sampling and mapping. The core idea is to map any arbitrary 6D unit direction to a point on the GWB. To achieve this, we begin by defining a support mapping $s_\mathcal{A}$ for a nonempty compact set $\mathcal{A}\subset\mathbb{R}^n$:
\begin{equation}
    \label{eq: support_mapping}
    s_\mathcal{A}(\mathbf{u}) = \underset{\mathbf{a}\in\mathcal{A}}{\arg \max}~\mathbf{u}^T\mathbf{a}, \|\mathbf{u}\|=1
\end{equation}
As illustrated in Fig. \ref{fig:gws}(1), $s_\mathcal{A}(\mathbf{u})$ yields the elements in $\mathcal{A}$ with the largest projection onto the given unit direction $\mathbf{u}\in\mathbb{R}^n$. This mapping also has four useful properties~\cite{lay2007convex, zheng2009improving}: 

\begin{property}
\label{prop_1}
    $\forall \mathbf{u}$, $s_\mathcal{A}(\mathbf{u}) \in \partial\mathcal{A}$.
\end{property}

\begin{property}
\label{prop_convex}
    If $\mathcal{A}$ is convex, $\forall \mathbf{x}\in\partial\mathcal{A}$, $\exists \mathbf{u}$, s.t. $s_\mathcal{A}(\mathbf{u})=\mathbf{x}$.
\end{property}

\begin{property}
\label{prop_add}
$s_\mathcal{A\bigoplus B}(\mathbf{u}) = s_\mathcal{A}(\mathbf{u})+s_\mathcal{B}(\mathbf{u})$.
\end{property}

\begin{property}
\label{prop_mul}
$s_{\mathbf{C}(\mathcal{A})}(\mathbf{u}) = \mathbf{C}\cdot s_\mathcal{A}(\mathbf{C}^T \mathbf{u})$, where $\mathbf{C}\in \mathbb{R}^{p\times n}$ and $p$ is any positive integer. (proved in the appendix)
\end{property}

Here, Prop. \ref{prop_1} indicates that the mapping $s_\mathcal{A}(\mathbf{u})$ always yields an element on $\partial\mathcal{A}$, which is the boundary of the set $\mathcal{A}$. Prop. \ref{prop_convex} shows that if the set $\mathcal{A}$ is convex, then $s_\mathcal{A}(\mathbf{u})$ is a surjection, indicating that any point on $\partial\mathcal{A}$ can be mapped. In our scenario, where $\mathcal{F}_i$, $\mathcal{W}_i$, and $\mathcal{W}_g$ are all convex, every point on the GWB can be mapped by $s_{\mathcal{W}_g}$. 
Despite the complexity of $s_{\mathcal{W}_g}$, we can simplify it by Prop. \ref{prop_add} and \ref{prop_mul}:
\begin{equation}
    \label{eq: swu}
    s_{\mathcal{W}_g}(\mathbf{u})
    =s_{\bigoplus_{i=1}^{m}\mathcal{W}_i}(\mathbf{u})
    =\sum_{i=1}^{m}s_{\mathcal{W}_i}(\mathbf{u})
    =\sum_{i=1}^{m}\mathbf{G}_i s_{\mathcal{F}_i}(\mathbf{G}_i^T \mathbf{u}).
\end{equation}
where $\bigoplus$ is the Minkowski sum and $m$ is the contact number.

If we acquire $s_{\mathcal{F}_i}$, we can readily deduce $s_{\mathcal{W}_g}$. In the context of the PCF contact model, each $\mathcal{F}_i$ represents a 3D cone. Since our GWS is defined under the $L_\infty$ assumption, $\mathcal{F}_i$ becomes a finite cone with a height of 1. We can use geometric intuition to directly determine $s_{\mathcal{F}_i}$:
\begin{equation}
    \label{eq: sfu}
    s_\mathcal{F}(\mathbf{u}) = \left\{
    \begin{array}{ll}
        \{\mathbf{f}~|~f_1=1, \mathbf{f}\in\mathcal{F}\}, & \text{if}~\theta=0\\
        \mathbf{v} & \text{if}~0 < \theta < \alpha  \\
        \{k\mathbf{v}~|~0 \le k \le 1\}, & \text{if}~\theta = \alpha \\
        \mathbf{o}, & \text{if}~\alpha < \theta \le \pi \\
    \end{array} \right.
\end{equation}
As shown in Fig. \ref{fig:gws}(2), $\mathbf{v}$ represents the intersection between the circle $\{\mathbf{f}~|~f_1=1, f_2^2+f_3^2=\mu^2\}$ and the plane spanned by $\mathbf{c}$ and $\mathbf{u}$, where $\mu$ is the friction coefficient, and $\mathbf{c}=(1, 0, 0)$. We also denote $\theta=\angle (\mathbf{c}, \mathbf{u})$ and $\alpha=\angle (\mathbf{c}, \mathbf{v}) + \frac{\pi}{2}=\arctan(\mu)+\frac{\pi}{2}$. 

Combining Eq. \ref{eq: swu} and \ref{eq: sfu}, we achieve the desired mapping from any arbitrary 6D unit direction to points on the GWB without any approximation. To obtain dense points $\{\mathbf{w}_k\}~(k=1,2,...,K)$ on $\partial\mathcal{W}_g$, we only need to uniformly sample $K$ ($K$ is a hyperparameter) $\mathbf{u}_k$ and map them using $s_{\mathcal{W}_g}$. This process does not involve optimization and can be fully paralleled for different contacts and samples. Additionally, this approach reduces the exponential time complexity of the classic discretization-based method~\cite{ferrari1992planning} for the contact number to linear complexity.

\textbf{Differentiability.} $s_{\mathcal{W}_g}(\mathbf{u})$ is differentiable with respect to the contact position $\mathbf{p}_i$ and normal $\mathbf{n}_i$ when $s_{\mathcal{F}_i}(\mathbf{G}_i^T\mathbf{u})\neq \mathbf{0}$. This is because $s_{\mathcal{W}_g}(\mathbf{u})$ in Eq. \ref{eq: swu} is differentiable with respect to the matrix $\mathbf{G}_i$ when $s_{\mathcal{F}_i}(\mathbf{G}_i^T\mathbf{u})\neq \mathbf{0}$. Moreover, the matrix $\mathbf{G}_i$ in Eq. \ref{eq: G_pcf} is differentiable with respect to the contact position $\mathbf{p}_i$ and normal $\mathbf{n}_i$.

\begin{figure}
    \centering
    \includegraphics[width=\linewidth]{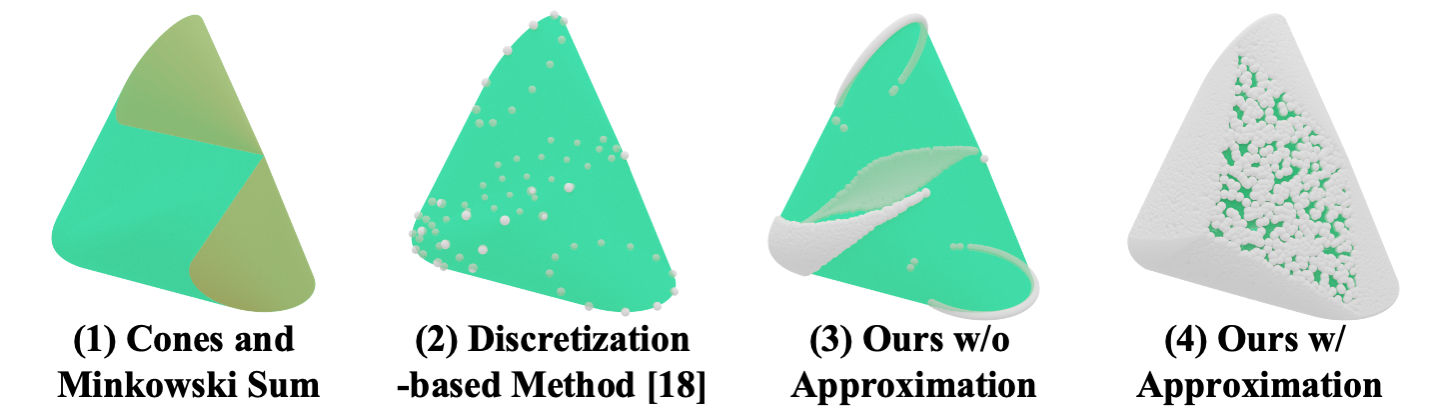}
    \caption{\textbf{GWS estimation visualized in 3D force space.} (1) In this example, GWS is the Minkowski sum of two cones. (2,3,4) Points (white) are sampled on GWB (green) by different methods. (4) Our method with approximation can get dense samples.}
    \label{fig:gws vis}
\end{figure}

\textbf{Approximation.} One limitation of the mapping $s_\mathcal{F}$ in Eq. \ref{eq: sfu} is that, when we randomly sample the unit vector $\mathbf{u}$, there is an extremely low probability that it will result in $\theta$ being either 0 or $\alpha$. As a result, certain areas on the GWB cannot be mapped, as illustrated in Fig. \ref{fig:gws vis}(3). To alleviate this issue, we propose to use an approximation:
\begin{equation}
    \label{eq: approx}
    s_\mathcal{F}(\mathbf{u}) = \left\{
    \begin{array}{ll}
        \mathbf{c}, &\text{if}~\theta=0\\
        \mathbf{c} + \theta / \delta \cdot(\mathbf{v}-\mathbf{c}), &\text{if}~ 0 < \theta < \delta \\
        \mathbf{v}, &\text{if}~ \delta \leq \theta \leq \alpha - \delta \\
        (\alpha-\theta) / \delta \cdot \mathbf{v}, &\text{if}~ \alpha - \delta  < \theta < \alpha  \\
        \mathbf{o}, &\text{if}~ \alpha \leq \theta \leq \pi \\
    \end{array} \right.
\end{equation}
Here, $\delta$ is a hyperparameter that governs the extent of the approximation. 
The basic idea is to loosen the equality conditions and interpolate between neighboring cases. Notably, we avoid relaxing around the origin $\mathbf{o}$ to prevent potential impacts on sampled points near the origin, maintaining metric accuracy. Because the $\epsilon$ metric can be computed as the smallest magnitude of sampled points on the GWS.

\textbf{Contact Position Normalization.} 
Another challenge is the variability in the L2 norm of the contact position $\mathbf{p}_i$. As the L2 norm of the contact normal $\mathbf{n}_i$ is always 1, if $\|\mathbf{p}_i\|$ is significantly smaller than $\|\mathbf{n}_i\|$, the torque space's magnitude becomes much smaller than that of the force space. In such cases, the GWS is a 6D flat ellipsoid. However, only when the GWS is a 6D sphere can evenly distributed $\mathbf{u}$ be mapped to evenly distributed points on the GWS. If the GWS deviates too much from a sphere, the sampled points on the GWS will be unevenly distributed, affecting grasp synthesis.

To alleviate this problem and achieve frame invariance,
we employ the transformation $\mathbf{p}'_i=\frac{1}{d}(\mathbf{p}_i-\bar{\mathbf{p}})$, similar to~\cite{zheng2009improving}, to normalize the contact position $\mathbf{p}_i$ before Eq. \ref{eq: G_pcf}. Here, $\bar{\mathbf{p}}=\frac{1}{m}\sum_{i=1}^m \mathbf{p}_i$ and $d=\frac{1}{m}\sum_{i=1}^m \|\mathbf{p}_i-\bar{\mathbf{p}}\|_2$.

\subsection{Differentiable Task-Oriented Energy}
\label{sec: task_loss}

\begin{figure}[t]
  \centering
  \includegraphics[width=0.8\linewidth]{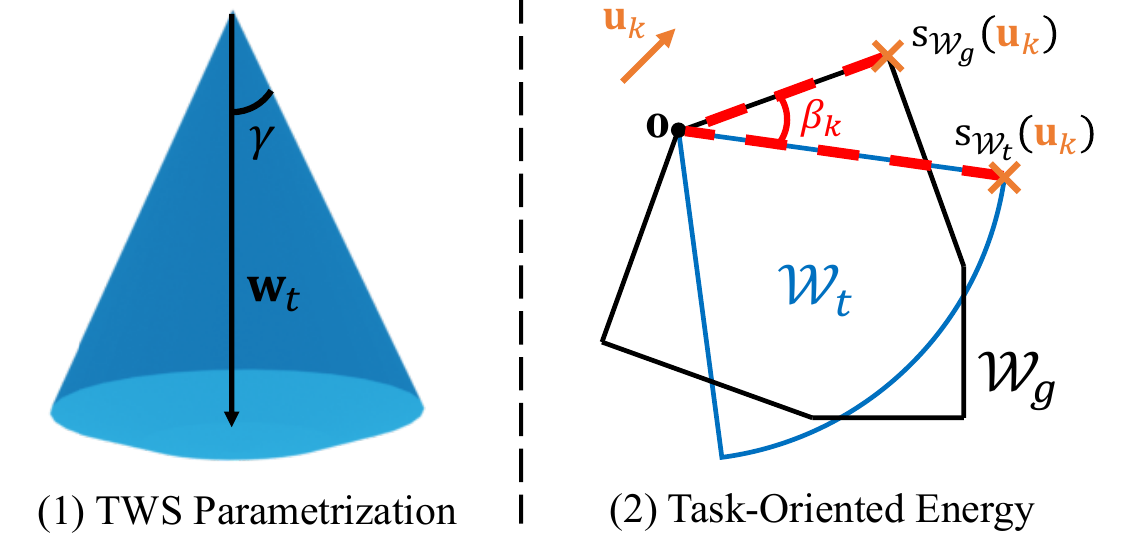}
  \caption{\textbf{Task-oriented energy.} (1) TWS is formulated as a 6D hyper-spherical sector parametrized by a 6D unit vector $\mathbf{w}_t$ and an angle $\gamma$. (2) The task-oriented energy is the sum of the cosine distance between $s_{\mathcal{W}_t}(\mathbf{u}_k)$ and $s_{\mathcal{W}_g}(\mathbf{u}_k)$. }
  \label{fig:loss}
\end{figure}

As illustrated in Fig. \ref{fig:loss}, the TWS is formulated as a 6-dimensional hyper-spherical sector:
\begin{equation}
    \label{eq: tws_def}
    \mathcal{W}_t(\mathbf{w}_t, \gamma)=\{\mathbf{t}\in\mathbb{R}^6~|~\angle(\mathbf{t}, \mathbf{w}_t) \leq\gamma, \|\mathbf{t}\|=1 \}
\end{equation}
Here, $\mathbf{w}_t\in\mathbb{R}^6$ can represent either the average external perturbations to resist or the average active wrenches to apply.
Moreover, the angle $\gamma\in \mathbb{R}$ can denote the tolerance to the potential disturbances, estimation errors, and external force fluctuations during task execution. A larger $\gamma$ implies a more robust grasp. When $\gamma=\pi$, the TWS becomes a hypersphere, representing the force closure task. 
In this work, we do not delve into the optimal choice of $\mathbf{w}_t$ and $\gamma$ for each task~\cite{lin2015grasp}, and we assume they are given as task priors.

For the energy design, optimizing the metric $\epsilon_t$ in Eq. \ref{eq:epsilon} directly would be ideal. However, $\epsilon_t$ is not differentiable when a wrench direction in TWS is not covered by GWS (\textit{i.e.,} $\epsilon_t=0$). This non-differentiability arises because $s_{\mathcal{W}_g}(\mathbf{u})$ lacks differentiability when $s_{\mathcal{W}_g}(\mathbf{u})=\mathbf{0}$. Additionally, when initializing a hand pose randomly, it's highly probable to have some wrench directions in TWS not covered by GWS, making $\epsilon_t$ unsuitable for optimization.

To enable optimization, irrespective of whether $\epsilon_t=0$, we propose a novel energy based on our GWS estimator. For each randomly sampled $\mathbf{u}_k$, the same mapping as in Eq. \ref{eq: support_mapping} is used to calculate an element $s_{\mathcal{W}_t}(\mathbf{u}_k)$ on TWS. Since our TWS is a hyper-spherical sector, solving $s_{\mathcal{W}_t}(\mathbf{u})$ is straightforward. As shown in Fig. \ref{fig:loss}, our energy is:
\begin{equation}  
    \label{eq: tws_loss}
    E_{t} = - \sum_{k=1}^K \cos{\beta_k} 
\end{equation}
where $\beta_k=\angle (s_{\mathcal{W}_t}(\mathbf{u}_k), s_{\mathcal{W}_g}(\mathbf{u}_k))$. $K$ is the sample number.

The cosine distance-based energy term $E_t$ strongly correlates with the metric $\epsilon_t$ introduced in Eq. \ref{eq:epsilon}. This connection arises from the shared behavior of $E_t$ and $\epsilon_t$. They both reach their minimum values only when the geometric shape of the GWS precisely aligns with that of the TWS. This similarity is the reason why we choose cosine distance over L2 distance. The L2 distance is influenced by the magnitudes of $s_{\mathcal{W}_g}(\mathbf{u}_k)$, potentially altering the minimum point. Using cosine distance ensures a more robust alignment based on geometric similarity rather than magnitude considerations.

\subsection{Dexterous Grasp Synthesis Pipeline}
\label{sec: hand synthesis}
Our dexterous grasp synthesis pipeline is built upon cuRobo~\cite{curobo_report23}. Our pipeline takes as input an object mesh $O$, task parameters $\{\mathbf{w}_t, \gamma\}$, and the expected hand contact points $\{\mathbf{x}^i_{rest}\}_{i=1}^{m}$ in the rest hand pose ($m$ is the contact number). Unlike DexGraspNet~\cite{wang2023dexgraspnet}, our hand contact points $\{\mathbf{x}^i_{rest}\}$ remain unchanged during optimization. The output variable to optimize is the hand pose $\mathbf{q}$, including root rotation, translation, and joint angles. For optimization, we employ gradient descent with greedy line search.

During optimization, the hand pose $\mathbf{q}$ is used to calculate the transformation of each hand link via forward kinematics. These transformations are then applied to $\{\mathbf{x}^i_{rest}\}_{i=1}^{m}$ to obtain the posed hand contact points $\{\mathbf{x}^i\}_{i=1}^{m}$. Next, these points are utilized to calculate the contact points $\{\mathbf{p}^i\}$ and normals $\{\mathbf{n}^i\}$ by finding the nearest points on the object mesh. To approximate the derivation of the nearest-point calculation, we use finite differences.

The optimization has a total of four energy terms. The first one is the task-oriented energy in Section \ref{sec: task_loss}, computed using the contact points $\{\mathbf{p}^i\}$ and normals $\{\mathbf{n}^i\}$. The second is the distance energy $E_d=\sum_{i=1}^{m}\|\mathbf{x}^i-\mathbf{p}^i\|^2$, aiming to encourage the hand to make contact with the object. Additionally, we incorporate the penetration and self-penetration energies (denoted as $E_p$) from cuRobo to prevent collisions between the hand and the object.

\section{Experiments}

This section begins with a sanity check for our proposed GWS estimator. Next, we present quantitative and qualitative results for task-oriented hand pose synthesis across 10 tasks. We also deploy some synthesized poses in the real world. Finally, we compare with DexGraspNet~\cite{wang2023dexgraspnet} to show our superiority in large-scale force closure grasp synthesis.

\subsection{Sanity Check for Grasp Wrench Space Estimation}

\textbf{Experiment setup.} 
To validate the efficiency of our GWS estimator, we experiment with 3 parameters: the number of contact points $m$, approximation angle $\delta$, and sampling number $K$. For each parameter setting, we randomly sample $m$ contact points on 49 objects from YCB dataset~\cite{xiang2017posecnn}, and use them to calculate GWS. We also experiment with different friction coefficients $\mu = [0.2, 0.3, 0.5, 1.0]$, resulting in $49 \times 4=196$ test cases per experiment. All metrics below are averaged over these 196 test cases.

We use the classic discretization-based method~\cite{ferrari1992planning} under $L_\infty$ assumption as our baseline and experiment with the discretization number $d=4,6,8$. Another two parameters $\delta$ and $K$ in our method are set as $15^\circ$ and $1e6$, respectively. 

\textbf{Metrics.} We evaluate the accuracy, density, and speed by three metrics: 1) \textbf{average relative length error} (RLE) (unit: $10^{-2}$) of sampled points $\mathbf{w}$ to the ground truth boundary in the direction of $\mathbf{w}$. The error of each point is $\max_{q\mathbf{w}\in\partial\mathcal{W}_g^{gt}} \frac{q-1}{q}$, which can be viewed as a Second-order Cone Program (SOCP) and solved by \cite{Domahidi2013ecos}. 
2) \textbf{Sparsity} (SP) (unit: rad), the expectation of the min angle between sampled points $\mathbf{w}$ and uniformly distributed points on the 6D sphere. Since this metric requires GWS to fully cover the sphere, we filter out non-force-closure test cases by randomly sampling many 6-subsets of $\{\mathbf{w}^{k}\}$ as simplices and checking whether all orthogonal wrench bases can be expressed as non-negative linear combinations of vertices of a simplex. We resample contact points until we have 196 test cases. 3) average \textbf{time} (t) (unit: ms) for each test case.

\begin{table}[t]
    \centering
    
    \begin{tabular}{|c|c|c|c|c|c|c|c|c|}
    \hline 
        & \multicolumn{4}{c|}{5 Contacts} & \multicolumn{4}{c|}{7 Contacts} \\
    \hline 
       \multirow{2}*{ }  & \multicolumn{3}{c|}
         {Baseline} & \multirow{2}*{Ours} & \multicolumn{3}{c|}
         {Baseline} & \multirow{2}*{Ours} \\
    \cline{2-4}
    \cline{6-8}
       & 4 & 6 & 8 &  & 4 & 6 & 8 & \\
    \hline 
        RLE$\downarrow$ &  5.30 & 2.36 & 1.26 & \textbf{0.43} & 6.49 & 2.78 & - & \textbf{0.70} \\
    \hline 
        SP $\downarrow$ &  0.48 & 0.42 & 0.38 & \textbf{0.29} & 0.36 & 0.31 & - & \textbf{0.26} \\
    \hline 
        t $\downarrow$ & 4e3 & 2e4 & 4e4 & \textbf{20} & 5e4 & 2e5 & 2e6 & \textbf{20} \\
    \hline 
        
    \end{tabular}
    \caption{\textbf{Sanity check for GWS estimator.} '-' indicates that we did not run it, since it is extremely slow (about 0.5 hours per case).}
    \label{tab:gws baseline}
    
\end{table}

\textbf{Main results.} Table \ref{tab:gws baseline} demonstrates our superior speed, accuracy, and density compared to the discretization-based baseline. The great speed advantage arises because our method can parallelize on GPU without any coding trick, whereas their QuickHull algorithm (Pyhull package) has to iterate on CPU. As the contact number $m$ increases, our time increases linearly, whereas theirs increases exponentially. Additionally, our method exhibits denser and more accurate results, because our method can map to any point in the GWB with fewer approximations.

\textbf{Hyperparameter analysis}.
In Tab. \ref{tab: ablation}, we investigate the effect of $\delta$ and $K$ on the quality of estimated GWS with 5 contact points. A $\delta$ of $0^\circ$ results in accurately sampled points on the GWS, but with uneven distribution. $\delta=15^\circ$ is a good choice to balance accuracy and uniformity. Higher $K$ boosts the density but increases computation time. 

\begin{table}[t]
\begin{center}
\begin{tabular}{|c|c|c|c|c|c|c|c|c|}
\hline
\multirow{2}*{} & \multicolumn{4}{c|}{$\delta$ (with $K=1e5$)} & \multicolumn{4}{c|}{K (with $\delta=15^\circ$)} \\
\cline{2-9}
& 0$^\circ$ & 15$^\circ$ & 30$^\circ$ & 45$^\circ$ & 1e3 & 1e4 & 1e5 & 1e6 \\
\hline
RLE$\downarrow$ & \textbf{0.00} & 0.42 & 5.45 & 19.4 & 0.43 & \textbf{0.42} & \textbf{0.42} & 0.43 \\
\hline
SP $\downarrow$ & 0.44 & \textbf{0.36} & \textbf{0.36} & \textbf{0.36} & 0.55 & 0.45 & 0.36 & \textbf{0.29} \\
\hline
t $\downarrow$ & \textbf{3.2} & \textbf{3.2} & \textbf{3.2} & \textbf{3.2} & \textbf{1.7} & 1.9 & 3.2 & 19.4 \\
\hline
\end{tabular}
\end{center}
\caption{\textbf{Hyperparameter analysis} on the quality of estimated GWS with 5 contact points.}
\label{tab: ablation}

\end{table}

\begin{table*}[t]
    \centering
    \begin{tabular}{|c|c|c|c||c|c|c|c|c|c|c|c|c|c|c|}
    \hline
       & \multirow{2}{*}{$E_d, E_p$} & \multirow{2}{*}{$E_t$} & \multirow{2}{*}{CPN} & \multicolumn{3}{c|}{Turn} & \multicolumn{2}{c|}{Lift} & \multicolumn{2}{c|}{Push/Pull} & \multicolumn{3}{c|}{Precision grasp} & \multirow{2}{*}{Average} \\
    \cline{5-14}
        & & & & Handle & Knob2 & Knob3 & Handbag & Plate & Button & Drawer & Lid & Key & USBA & \\
    \hline
        Baseline& \checkmark & & &13	&39&	44	&0&	25	&40	&10&	38	&27&	2	&23.8
 \\
     \hline
        Ablation1 & \checkmark & L2 & \checkmark &\textbf{84}&	60	&64	&10	&80	&91&	48	&\textbf{97}	&\textbf{68}	&\textbf{62}	&66.4
\\
     \hline
        Ablation2 & \checkmark & cos & & 79	&64	& \textbf{72}	&50&24&	96	&\textbf{99}	&86	&19	&2	&59.1 \\
     \hline
        Ours & \checkmark & cos & \checkmark & 75	&\textbf{73}	&58	&\textbf{51}	&\textbf{97}	&\textbf{97}	&\textbf{99}	&74	&62	&40	& \textbf{72.6}
\\
    \hline
        
    \end{tabular}
    \caption{\textbf{Simulation success rate ($\%$) for task-oriented pose synthesis.} Three variants of our method consistently outperform the baseline. Our final version performs equally well on different tasks, while other variants fail in some cases.}
    \label{tab: task oriented 1}
\end{table*}

\begin{figure*}
    \centering
    \includegraphics[width=\linewidth]{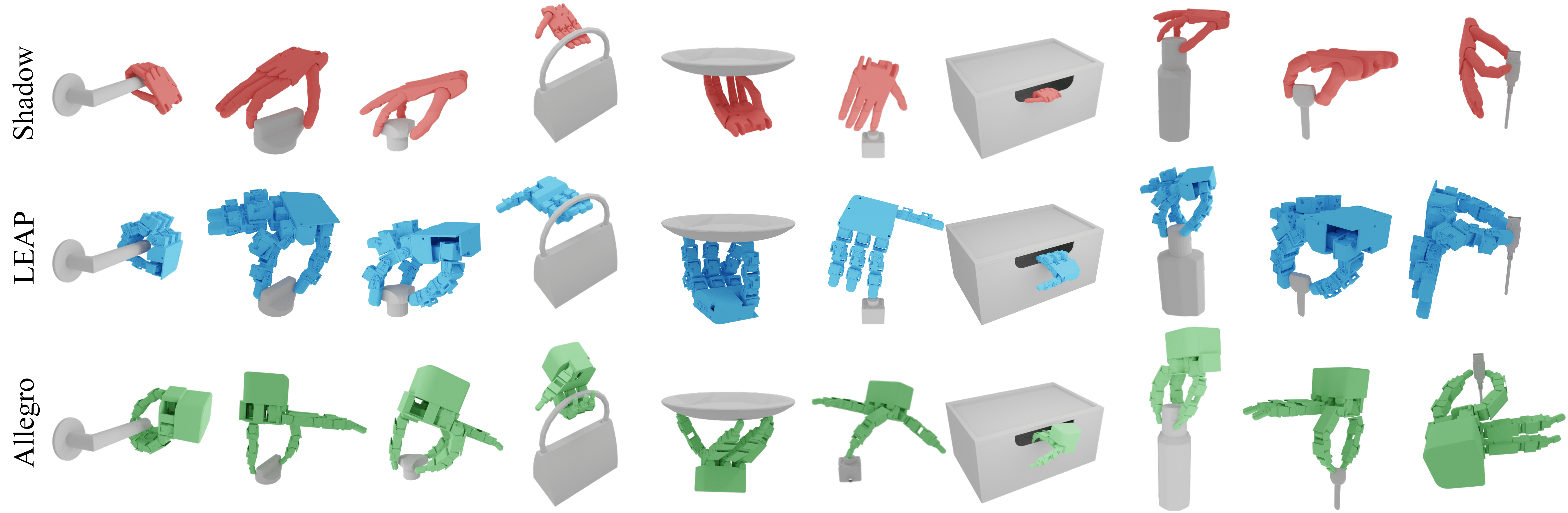}
    \caption{\textbf{Visualization of synthesized task-oriented poses.} We show similar poses for the Shadow hand and the LEAP hand, but different poses for the Allegro hand.}
    \label{fig:gallery}
\end{figure*}

\subsection{Task-Oriented Dexterous Hand Pose Synthesis}
\label{sec:task oriented exp}

\textbf{Experiment setup.} As there is no standard benchmark for evaluating task-oriented dexterous grasp synthesis, we gathered 10 different objects and created 10 unique tasks. These tasks involve turning, lifting, pushing, pulling, and precision grasping. We optimize 500 iterations and generate 100 grasps per task. 

For each task, three conditions must be specified: Task Wrench Space (TWS), hand contact points, and hand pose initialization. Take the task of screwing a knob as an example, we define the TWS parameters as $\mathbf{w}_t=[0,0,0,0,0,1]$ and $\gamma=15^\circ$. The hand contact points are set to be the farthest spheres in the distal link of the index and thumb fingers (cuRobo requires segmenting the robot into spheres). The initial hand position is set above the knob, with the palm facing downward. A random perturbation is then applied to the hand position and orientation. Initially, all hand joint angles are set to 0.

Based on the above example, we only modify some necessary parts for a different task. This usually involves modifying $\mathbf{w}_t$, determining which hand fingers to use, and setting the initial hand position and orientation. Specifically, for precision grasps, we set $\gamma=180^\circ$.

\textbf{Metrics.} 1) \textbf{Simulation success rate} (SS) (unit:\%) in Isaac Gym~\cite{makoviychuk2021isaac}. During evaluation, the hand pose is fixed, and an external wrench is applied to each object at each simulation step. The task succeeds if the object remains unmoved after simulating 100 steps. We conduct a single test for most tasks with the applied external wrench being negative to the $\mathbf{w}_t$ of the TWS. In the case of precision grasps, we follow the testing procedure for force-closure grasps in DexGraspNet~\cite{wang2023dexgraspnet}, conducting the test 6 times with 6 different force directions. For tasks involving turning, pushing, and pulling, we use a virtual revolute or prismatic joint for each object to restrict its movement. 2) \textbf{Time} (t) (unit: s) for synthesizing 100 grasps on an Nvidia RTX 3090 GPU card. 3) \textbf{Memory} (M) (unit: GB) cost on GPU.

\textbf{Main results.} In Table \ref{tab: task oriented 1}, we compare four methods: (1) \textit{Baseline}, which doesn't incorporate task-oriented energy. (2) \textit{Ablation1}, utilizing L2 distance as the task-oriented energy instead of cosine distance. (3) \textit{Ablation2}, excluding contact position normalization in Sec. \ref{sec:gws}. (4) \textit{Ours}. 

We can see that all three variants of our method consistently outperform the baseline, confirming the effectiveness of our task-oriented energy. Our final version performs equally well on different tasks, while other ablations sometimes fail. The L2 energy suffers from low success rates if the magnitude of the GWS significantly differs from the TWS. For example, when using 4 fingers to pull a drawer, the magnitude of GWS should be approximately 4 times larger than that of the TWS. Moreover, without CPN, the GWS estimator and the task-oriented energy will prioritize the force space over the torque space if the contact position has a much smaller magnitude than 1. This can significantly impact tasks that require consideration of both force and torque for success. For example, the plate cannot be held stably with a non-zero torque.

\begin{table}[]
    \centering
    \begin{tabular}{|c|c|c||c|c|c|c|}
    \hline
         & $\delta$& $K$&  SS ($\%$)$\uparrow$ & t ($s$) $\downarrow$ & M (GB) $\downarrow$ \\
     \hline
        Ours & $15^\circ$ & $1e2$ & 72.6 & 14.9 & \textbf{1.5} \\
    \hline
        Ablation3 &$15^\circ$ & $1e4$ & \textbf{75.1} & 83.5 & 15.3 \\
    \hline
      Ablation4 &$0^\circ$ & $1e2$  & 72.1 & \textbf{14.5} & \textbf{1.5} \\
   
    \hline
       Ablation5 & $0^\circ$ & $1e4$ & 74.8 & 83.4 & 15.3\\
    \hline
    \end{tabular}
    \caption{\textbf{Hyperparameter analysis} for task-oriented hand pose synthesis.}
    \label{tab:task oriented 2}
\end{table}

\textbf{Hyperparameter analysis.} In Table \ref{tab:task oriented 2}, we investigate the impact of two parameters, the approximation angle $\delta$ and the sampling number $K$, in the GWS estimator for grasp synthesis. The default values in Table \ref{tab: task oriented 1} are $\delta=15^\circ$ and $K=1e2$, striking a balance between speed and success rate. As shown in Table \ref{tab:task oriented 2}, a larger $K$ improves the average success rate but slows down the synthesis. Leveraging the approximation can slightly improve performance without incurring additional costs.

\begin{figure}[t]
    \centering
    \includegraphics[width=\linewidth]{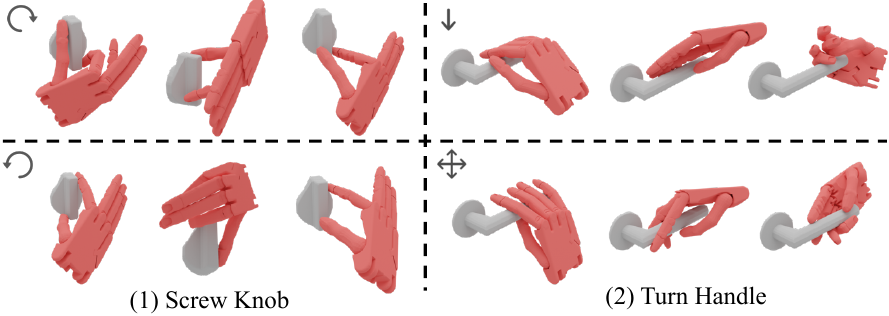}
    \caption{\textbf{Visualization of synthesized poses for different TWS.} Each TWS is denoted by the upper left arrow, while the fourth TWS means force closure. }
    \label{fig:task oriented grasp}
\end{figure}

\textbf{Visualization.} Finally, we visualize our synthesized grasps on each task for different robot hands in Fig. \ref{fig:gallery}, such as Shadow hand, LEAP hand~\cite{shaw2023leap}, and Allegro hand. Additionally, for knob screwing and handle turning, we change the TWS and show 3 diverse grasps per TWS in Fig. \ref{fig:task oriented grasp}. Each column shows that different TWS can synthesize different grasps with the same hand pose initialization. Each third grasp is an unnatural example, but the desired force/torque can still be applied.

\subsection{Real World Experiment for Task-Oriented Poses}

Next, we deploy some synthesized task-oriented hand poses in the real world for manipulation. The hardware setup includes a LEAP hand~\cite{shaw2023leap} mounted on a UR-5e arm and a Kinect V2 as the depth sensor. The object mesh is obtained by scanning, and its pose is estimated using the Iterative Closest Point (ICP) algorithm. To reach the synthesized hand poses, we use the original cuRobo library for motion planning. The following movements to manipulate the object are manually designed for each task. The design principle is to freeze the hand joints and only move the hand root. 

\begin{figure}[t]
    \centering
    \includegraphics[width=\columnwidth]{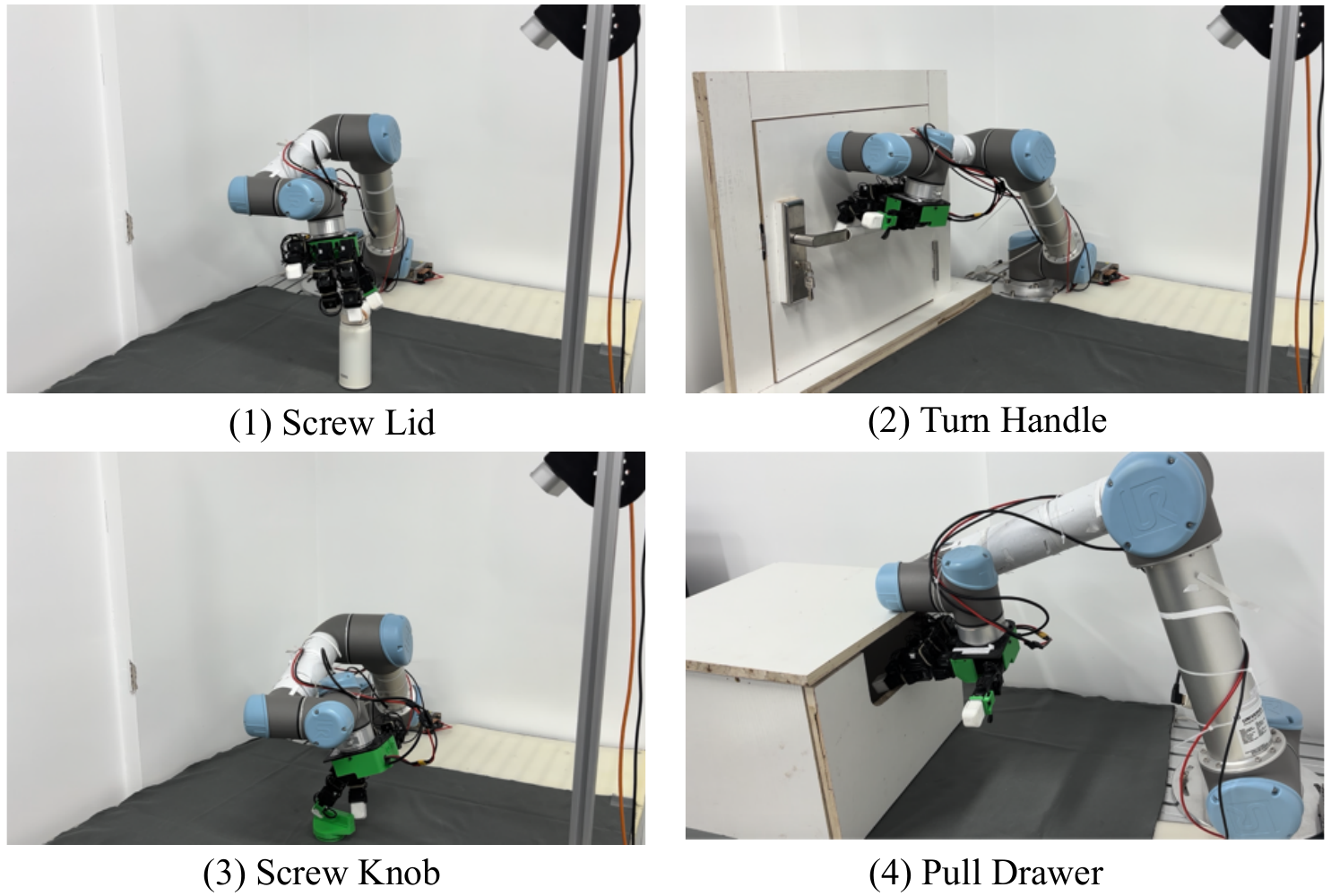}
    \caption{\textbf{Real-world deployments} across 4 tasks.}
    \label{fig: real world}
\end{figure}

In Fig. \ref{fig: real world}, we show experiments on lid screwing, handle turning, knob screwing, and drawer pulling. Please refer to the supplementary material for the video. Each task is tested in 10 trials with different grasp poses, resulting in success rates of 5/10, 2/10, 6/10, and 8/10, respectively. The handle-turning task has a low success rate because the restoring force is large. We also observe several failure cases when the finger misses the object by a small margin. Future work can incorporate close-loop tactile feedback to help in.

\subsection{Large-Scale Force-Closure Dexterous Grasp Synthesis}
\label{sec: fc exp}

\textbf{Experiment setup.} To evaluate the synthesis of force-closure dexterous grasps, we use the large-scale object assets from DexGraspNet~\cite{wang2023dexgraspnet}, comprising over 5700 objects. We optimize 500 iterations and generate 20 grasps for each object, resulting in more than 100,000 grasps. The baseline is established using the default settings of DexGraspNet. In our approach, we set $m=5$, $\delta=15^\circ$, and $K=10^2$ to balance the speed and performance. The hand contact points are designated as the farthest spheres in the distal link of each finger. To initialize the hand root pose, we adopt the same approach as DexGraspNet, positioning the hand at a random distance from the object with the palm facing toward it. 

 \textbf{Metrics.} 1) \textbf{Simulation success rate} (SS), similar to the one described in Section \ref{sec:task oriented exp}. 2)  \textbf{Max penetration depth} (MP) of each grasp. 3) $\epsilon$ \textbf{metric}. Using our GWS estimator, this metric can be calculated as the minimum magnitude of sampled points on GWB. The input to the GWS estimator, contact points, are chosen as the nearest points to the object on each hand link in contact. A link is considered in contact if the distance between a hand link and the object is smaller than 5$mm$. 4) \textbf{times} to synthesize all grasps on a GPU.

\textbf{Main results.} As shown in Tab. \ref{tab:fc_grasp}, our method can synthesize force-closure grasps significantly faster than DexGraspNet while maintaining comparable quality. This 50x acceleration can be attributed to two primary factors: First, we optimize for only 500 iterations, which is $10\times$ fewer than DexGraspNet, yet still yields satisfactory results, thanks to our good objective function. Second, we take advantage of the efficient CUDA functions in the cuRobo library. 

We can also find that DexGraspNet synthesizes grasps with a larger $\epsilon$ metric. This is mainly because their synthesized hands are often closer to the object, leading to more hand links being considered in contact. However, this proximity comes with the drawback of increased penetration compared to our approach.

\begin{table}[t]
\begin{center}
\begin{tabular}{|c|c|c|c|c|}
\hline
 & SS ($\%$) $\uparrow$ & MP ($mm$) $\downarrow$ & $\epsilon \uparrow$ & t (hour) $\downarrow$\\
\hline
DexGraspNet & 37.0 & 7.9 & \textbf{0.67} & 60.0 \\
\hline
Ours & \textbf{42.5} & \textbf{4.8} & 0.50 & \textbf{1.2} \\
\hline
\end{tabular}
\end{center}
\caption{\textbf{100k force-closure dexterous grasp synthesis.}}
\label{tab:fc_grasp}

\end{table}

\section{Limitations}

First, the current formulation can only synthesize a static pose but not a whole trajectory for complex manipulation tasks, \textit{e.g.}, in-hand reorientation. Second, the contact between the object and the environment is not modeled. Finally, employing a geometric primitive hyper-spherical sector to parameterize TWS may not be suitable for all tasks. Future research could explore the use of our method on more complex and long-horizon manipulation tasks.


\section{Conclusions}

This work proposes a unified framework for efficient task-oriented dexterous hand pose synthesis without human data. Our first contribution is a novel, fast, accurate, and differentiable approach to estimating the GWS. Based on it, we propose a novel task-oriented energy for optimizing dexterous hand poses for various tasks, including non-force-closure grasps, force-closure grasps, and non-prehensile manipulations. Extensive experiments verify the efficiency and effectiveness of our novel GWS estimator, task-oriented energy, and the improved synthesis pipeline.





\bibliographystyle{IEEEtran}
\bibliography{IEEEabrv}

\appendix

\subsection{Proof for Property \ref{prop_mul}}

\begin{proof}
    $s_{\mathbf{C}(\mathcal{A})}(\mathbf{u}) 
     = \underset{\mathbf{C}\mathbf{a}\in\mathbf{C}(\mathcal{A})}{\arg \max}~\mathbf{u}^T(\mathbf{C}\mathbf{a})\\ 
     ~~~~~~~~~~~~~~~~~~~~~~~~~~~= \mathbf{C}\cdot \underset{\mathbf{a}\in\mathcal{A}}{\arg \max}~(\mathbf{C}^T\mathbf{u})^T\mathbf{a} \\
     ~~~~~~~~~~~~~~~~~~~~~~~~~~~= \mathbf{C}\cdot s_\mathcal{A}(\mathbf{C}^T \mathbf{u})
    $
\end{proof}

\subsection{GWS Estimator for Soft Contact Model}

The $\mathcal{F}_i$ and $\mathbf{G}_i$ for the soft contact model (SFC) is:
\begin{flalign}
\label{eq: F_sfc}
& \mathcal{F}_i^{SFC} = \left\{\mathbf{f}_i\in\mathbb{R}^4~|~0 \leq f_{i1}, \frac{f_{i2}^2+f_{i3}^2}{\mu_1^2}+\frac{f_{i4}^2}{\mu_2^2} \leq f_{i1}^2 \right\} 
\\
\label{eq: G_sfc}
& \mathbf{G}_i^{SFC} = 
\begin{bmatrix}
    \mathbf{n}_i & \mathbf{d}_i & \mathbf{e}_i & \mathbf{0}\\
    \mathbf{p}_i \times \mathbf{n}_i & 
    \mathbf{p}_i \times \mathbf{d}_i & 
    \mathbf{p}_i \times \mathbf{e}_i & 
    \mathbf{n}_i \\
\end{bmatrix} \in \mathbb{R}^{6\times4}
\end{flalign}

Eq. \ref{eq: sfu} can be extended to the soft contact model in 4D. Solving $\mathbf{v}$ in 4D is the main difficulty. The mathematical intuition behind solving $\mathbf{v}$ in 3D is the Cauchy-Schwarz inequality. Therefore, in 4D, we can have 
\begin{align*}
    \mathbf{u}^T \mathbf{f} 
    & =\sum_{k=1}^{4} u_k f_k \\
    & \leq u_1 f_1 + \sqrt{\left(u_2^2+u_3^2+\frac{u_4^2\mu_2^2}{\mu_1^2}\right)\left(f_2^2+f_3^2+\frac{f_4^2\mu_1^2}{\mu_2^2}\right)} \\
    & \leq \left(u_1 + \sqrt{u_2^2\mu_1^2+u_3^2\mu_1^2+u_4^2\mu_2^2}\right) f_1
\end{align*}
The first inequality is tight when 
\begin{equation}
    \label{eq:cond}
    \frac{f_2}{u_2}=\frac{f_3}{u_3}=\frac{f_4\mu_2^2}{u_4\mu_1^2}.
\end{equation}
$\mathbf{v}$ is then the interaction between 2D circle $\{\mathbf{f}~|~f_1=1, \frac{f_{2}^2+f_{3}^2}{\mu_1^2}+\frac{f_{4}^2}{\mu_2^2}=1\}$ and the space that satisfy Eq. \ref{eq:cond}.


\end{document}